\documentclass[sigconf,10pt,nonacm,authorversion]{acmart}

\usepackage{amsmath,amsfonts} 
\usepackage[ruled,vlined]{algorithm2e}
\usepackage{graphicx}
\usepackage{textcomp}
\usepackage{xcolor}
\usepackage{tabularx}
\usepackage{caption}
\usepackage{subcaption}
\usepackage[inline]{enumitem}
\newcommand{\Ch}{\checkmark}
\newcommand{\X}{$\times$}
\usepackage{bm}
\usepackage{bbm}
\usepackage{url}
\usepackage{amsthm}
\usepackage{wrapfig}
\usepackage{multicol}
\usepackage{multirow}

\SetCommentSty{mycommfont}

\usepackage{xspace}
\newcommand{\Dataset}{SensorQA\xspace}

\newcommand{\cmark}{\ding{51}}%
\newcommand{\xmark}{\ding{55}}%
\definecolor{myred}{RGB}{170,0,0} 
\definecolor{myblue}{RGB}{0,0,170} 
\definecolor{mygreen}{RGB}{0,100,0} 

\AtBeginDocument{%
  }

\setcopyright{acmlicensed}
\copyrightyear{2018}
\acmYear{2018}
\acmDOI{XXXXXXX.XXXXXXX}

\acmConference[Under review]{Make sure to enter the correct
  conference title from your rights confirmation emai}{}{}
\acmISBN{978-1-4503-XXXX-X/18/06}

\begin{document}

\title{\Dataset: A Question Answering Benchmark for Daily-Life Monitoring}




\thanks{\textsuperscript{*}Both authors contributed equally to this research.}

\author{Benjamin Reichman\textsuperscript{*}}
\email{bzr@gatech.edu}
\affiliation{%
  \institution{Georgia Institute of Technology}
  \city{Atlanta}
  \state{Georgia}
  \country{USA}
}

\author{Xiaofan Yu\textsuperscript{*}}
\email{x1yu@ucsd.edu}
\orcid{0000-0002-9638-6184}
\affiliation{%
  \institution{University of California San Diego}
  \city{La Jolla}
  \state{California}
  \country{USA}
}

\author{Lanxiang Hu}
\email{lah003@ucsd.edu}
\affiliation{%
  \institution{University of California San Diego}
  \city{La Jolla}
  \state{California}
  \country{USA}
}

\author{Jack Truxal}
\email{jtruxal6@gatech.edu}
\affiliation{%
  \institution{Georgia Institute of Technology}
  \city{Atlanta}
  \state{Georgia}
  \country{USA}
}

\author{Atishay Jain}
\email{atishay.jain@gatech.edu}
\affiliation{%
  \institution{Georgia Institute of Technology}
  \city{Atlanta}
  \state{Georgia}
  \country{USA}
}

\author{Rushil Chandrupatla}
\email{ruchandrupatla@ucsd.edu}
\affiliation{%
  \institution{University of California San Diego}
  \city{La Jolla}
  \state{California}
  \country{USA}
}

\author{Tajana \v{S}imuni\'{c} Rosing}
\email{tajana@ucsd.edu}
\orcid{0000-0002-6954-997X}
\affiliation{%
  \institution{University of California San Diego}
  \city{La Jolla}
  \state{California}
  \country{USA}
}

\author{Larry Heck}
\email{larryheck@gatech.edu}
\affiliation{%
  \institution{Georgia Institute of Technology}
  \city{Atlanta}
  \state{Georgia}
  \country{USA}
}

\renewcommand{\shortauthors}{}

\begin{abstract}
With the rapid growth in sensor data, effectively interpreting and interfacing with these data in a human-understandable way has become crucial.
While existing research primarily focuses on learning classification models, fewer studies have explored how end users can actively extract useful insights from sensor data, often hindered by the lack of a proper dataset. To address this gap, we introduce \Dataset, the first human-created question-answering (QA) dataset for long-term time-series sensor data for daily life monitoring. \Dataset is created by human workers and includes 5.6K diverse and practical queries that reflect genuine human interests, paired with accurate answers derived from sensor data. We further establish benchmarks for state-of-the-art AI models on this dataset and evaluate their performance on typical edge devices. Our results reveal a gap between current models and optimal QA performance and efficiency, highlighting the need for new contributions. The dataset and code are available at: \url{https://github.com/benjamin-reichman/SensorQA}.
\end{abstract}



\settopmatter{printacmref=false}


\maketitle

\section{Introduction}
\label{sec:intro}


In recent years, the number of connected Internet-of-Things (IoT) devices has grown exponentially, with an estimated 40 billion devices expected by 2030~\cite{iotconnection}. These devices generate vast amounts of sensor data, which, unlike text or video, are not easily interpretable by humans due to their raw and complex nature. While existing machine learning algorithms can classify sensor data into predefined categories~\cite{ouyang2022cosmo, ouyang2023harmony, xu2023practically, xu2023mesen, cao2024mmclip, weng2024large}, they fall short in providing an intuitive way for humans to interact with and extract meaningful insights from this data. For example, answering a question like ``How good was my work-life balance last week?'' is straightforward for humans, but current technologies require multiple steps: (1) selecting the appropriate sensor data, (2) understanding the difference between work and relaxing activities, and (3) researching online to understand what qualifies as a healthy work-life balance. In everyday life, people are more interested in gaining insights related to their health and well-being, rather than identifying specific activities at a given moment.

\textbf{Question Answering (QA)} is an ideal framework for modeling natural interactions between humans and sensors: users ask questions and receive accurate answers based on the sensor data.
While QA has been extensively studied in the language and vision domains~\cite{squad,boolq,triviaqa,medquad,vqav2,qian2024nuscenes,okvqav2}, few studies have explored sensor applications using available sensor data.
Early QA systems such as \textit{AI therapist}~\cite{nie2022conversational} and DeepSQA~\cite{xing2021deepsqa} focus on mental health diagnosis and human activity monitoring. 
However, their QA dataset is generated by template-based searches, which presents limited diversity and practical value.
The recent rise of Large Language Models (LLMs) offers the potential for handling more diverse queries and sophisticated reasoning, such as captioning IMU signals related to activity~\cite{moon2023anymal, han2024onellm} or analyzing smartwatch data for medical diagnosis~\cite{englhardt2024classification,kim2024health,yang2024drhouse}. However, these models are currently constrained to dealing with short durations of sensor data, typically around 10 seconds, or low-dimensional sensor data such as step counts per day.

In summary, the progress of QA interactions for sensing applications is limited by \textit{the absence of suitable datasets and benchmarks}. An ideal dataset would include diverse, practical samples, such as raw sensor signals of varying durations collected in real-world settings, along with rich QA pairs that align with users' genuine interests. To the best of the authors' knowledge, no such benchmark has been proposed.

\textbf{Our Contribution} In this work, we introduce \Dataset, the \textit{first} human-created dataset and benchmark for QA interactions between humans and long-term time-series sensor data. The creation of \Dataset emphasizes realistic QA scenarios that closely resemble everyday life. On the sensor data side, \Dataset focuses on daily activity monitoring using commonly available mobile devices for sensing, such as smartphones and smartwatches, with extensive sensor data collected from 60 users intermittently over a period of up to three months.
For the QA part, we visualize activity schedules and present these graphs to human workers on the Amazon Mechanical Turk (AMT) platform. Workers are tasked with generating questions based on their practical interests and writing down ground-truth answers according to the activity graphs. To encourage question diversity, we design multi-timescale activity graphs using 14 different activity label subsets.
As a result, \Dataset contains 5.6K QA pairs covering different sensor time scales from one day to multiple weeks, with six question categories and seven answer categories.

The contributions of this work are summarized as follows:
\begin{itemize}[noitemsep,topsep=0pt]
    \item We present \Dataset, a human-created QA benchmark with naturally collected sensor data and diverse QA pairs, aimed at real-world scenarios.
    \item We benchmark state-of-the-art baselines on \Dataset using typical edge devices, highlighting the challenges in QA accuracy and efficiency.
    \item We open-sourced \Dataset and our code to encourage further contributions in this area.
\end{itemize}

\vspace{-2mm}
\section{Related Work}
\label{sec:related-work}

\begin{figure}[tb]
  \centering
  \includegraphics[width=0.47\textwidth]{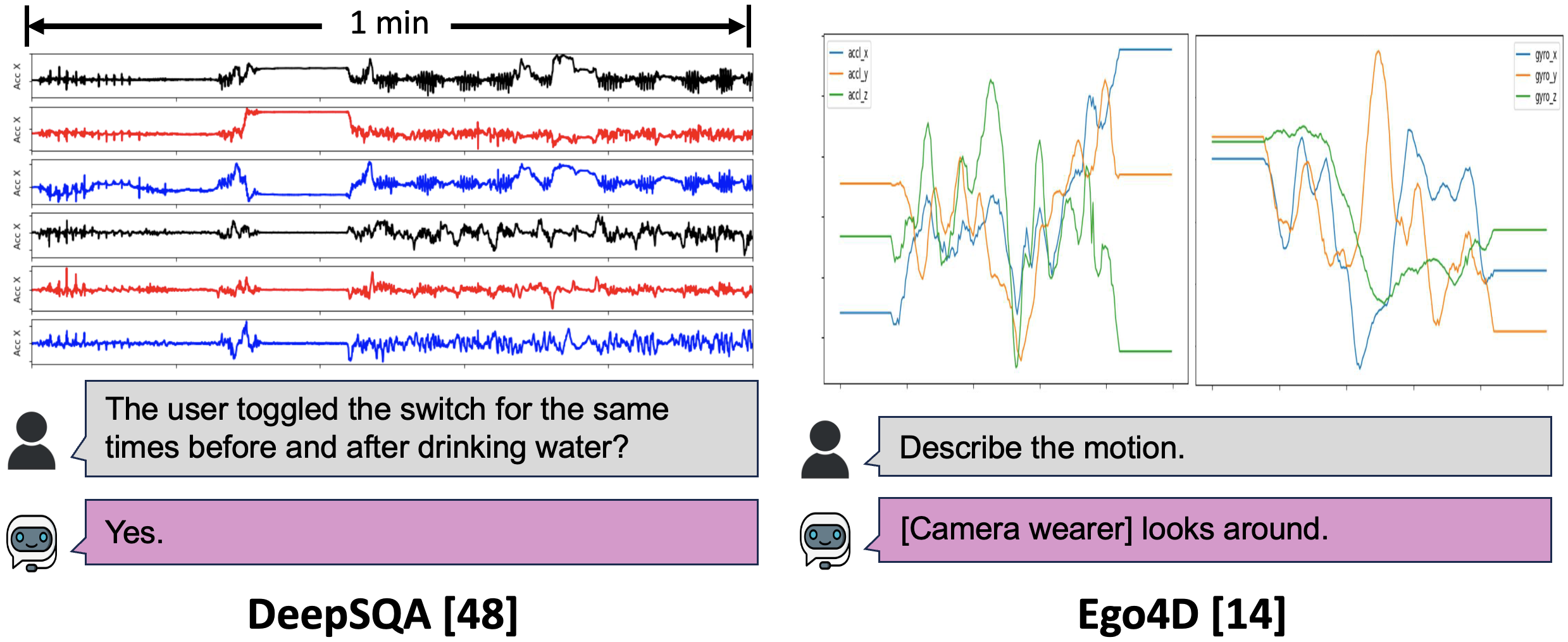}
  \vspace{-4mm}
  \caption{\small Visualizations of existing QA datasets using time series IMU sensor data.}
  \vspace{-5mm}
  \label{fig:data-compare2}
\end{figure}

\textbf{Question Answering using sensor data}
Question Answering has attracted extensive interests in the field of Natural Language Processing (NLP), with a wide range of datasets designed to address various language tasks~\cite{squad,boolq,triviaqa,medquad,jecqa,finqa,naturalquestions,hotpotqa,liu2024benchmarking,hu2024ket}. 
Recent benchmarks integrate multiple modalities in QA, such as VQAv2~\cite{vqav2} for visual question answering, ScienceQA~\cite{scienceqa} for science-related questions, and PathVQA~\cite{pathvqa} for pathology image-based questions, among others.
In sensing, researchers have developed multiple QA benchmarks for remote sensing~\cite{satbird,rsvqa,samvqa,earthvqa} and clinical diagnosis using low-dimensional sensor data~\cite{englhardt2024classification,kim2024health,yang2024drhouse}. 

DeepSQA~\cite{xing2021deepsqa} is currently the only QA dataset and benchmark for time-series data from IoT sensors, specifically for human activity monitoring. The dataset is generated automatically by a rule-based search algorithm. As a result, the questions in DeepSQA lack linguistic and content diversity, and, more importantly, may not reflect the practical interests of users. For example, as shown in Fig.~\ref{fig:data-compare2} (left), one predefined question template compares the frequency of activities at two different times, which may not provide meaningful insights in real-world scenarios.

\textbf{Multimodal reasoning using sensor data} 
Recent works have explored multimodal reasoning that connects sensor data with natural language. IMU2CLIP~\cite{moon-etal-2023-imu2clip}, mmCLIP~\cite{cao2024mmclip}, and TENT~\cite{zhou2023tent} align textual and sensor data embeddings using contrastive learning, similar to CLIP~\cite{radford2021learning}. FM-Fi~\cite{weng2024large} leverages vision-based models for radio-frequency sensing. The most relevant works, AnyMAL~\cite{moon2023anymal} and OneLLM~\cite{han2024onellm}, enable more advanced reasoning beyond activity classification. Both works connect sensor embeddings to an LLM via an adapter module, with the LLM fine-tuned on IMU data and text descriptions from the Ego4D dataset~\cite{grauman2022ego4d}, as shown in Fig.~\ref{fig:data-compare2} (right). However, all these methods are limited to simple reasoning tasks over short, fixed-duration signals.

\begin{table}[t]
\footnotesize
\caption{\small Comparing {\Dataset} and existing works using time series sensors.}
\label{tbl:related_works}
\vspace{-4mm}
\begin{center}
\begin{tabular}{ccc} 
\toprule
\textbf{Dataset/Benchmark} & \textbf{Human-created} & \textbf{Long-duration} \\
& \textbf{rich text} & \textbf{sensor data} \\
\midrule
DeepSQA~\cite{xing2021deepsqa} & \X & \X \\
AnyMAL~\cite{moon2023anymal}, OneLLM~\cite{han2024onellm} & \Ch & \X \\
 \textbf{\Dataset (this work)} & \Ch & \Ch \\
\bottomrule
\end{tabular}
\end{center}
\vspace{-4mm}
\end{table}

In summary, our \Dataset dataset significantly differs from prior benchmarks in two aspects:
(i) On the \textit{QA} side, \Dataset, created by humans, is highly diverse in terms of both the questions posed and the answers that are provided, better reflecting the needs of a user, (ii) On the \textit{sensor} side, \Dataset contains arbitrary length sensor signals and activity histories of up to multiple weeks. The comparison is detailed in Table~\ref{tbl:related_works}.

\begin{figure*}[htbp]
    \centering
    \begin{subfigure}[b]{0.48\textwidth}
        \centering
        \includegraphics[width=\textwidth]{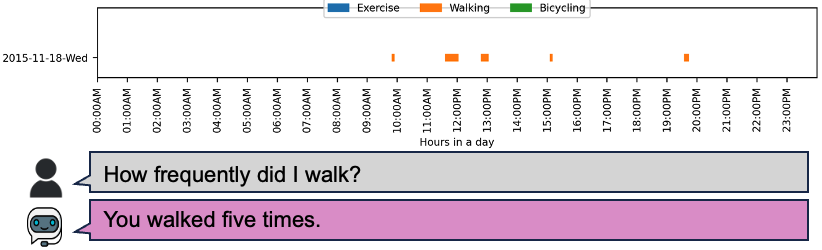}
        \vspace{-6mm}
        \caption{\small Example frequency query.}
        \label{fig:daily-freq}
    \end{subfigure}
    \hfill
    \begin{subfigure}[b]{0.48\textwidth}
        \centering
        \includegraphics[width=\textwidth]{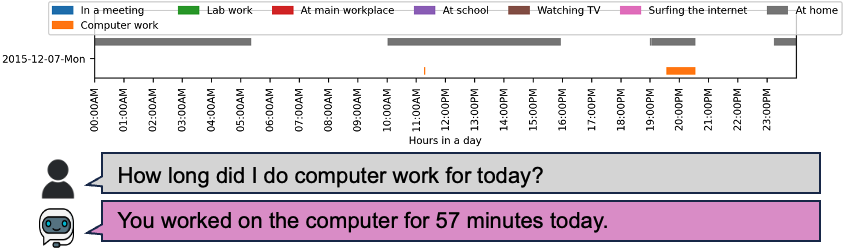}
        \vspace{-6mm}
        \caption{Example time query.}
        \label{fig:daily-time}
    \end{subfigure}
    
    \begin{subfigure}[b]{0.48\textwidth}
        \centering
        \includegraphics[width=\textwidth]{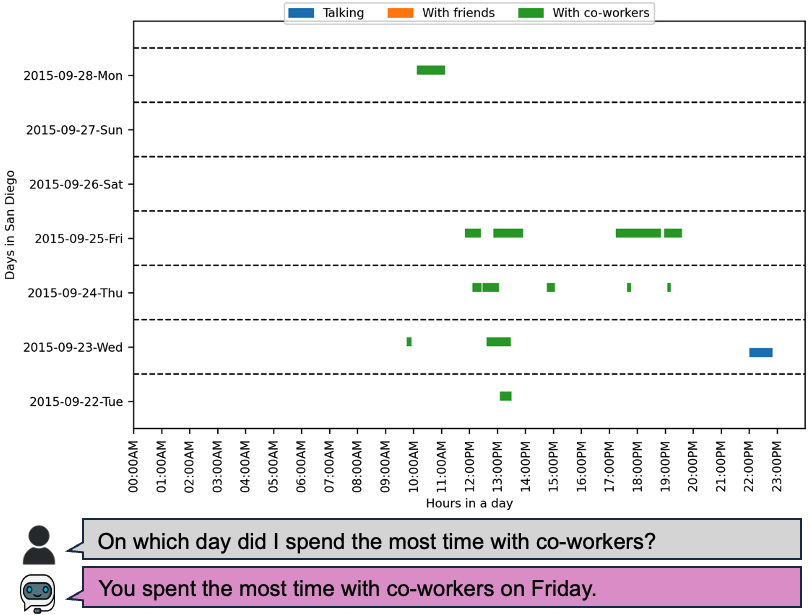}
        \vspace{-6mm}
        \caption{Example day query.}
        \label{fig:weekly-day}
    \end{subfigure}
    \hfill
    \begin{subfigure}[b]{0.48\textwidth}
        \centering
        \includegraphics[width=\textwidth]{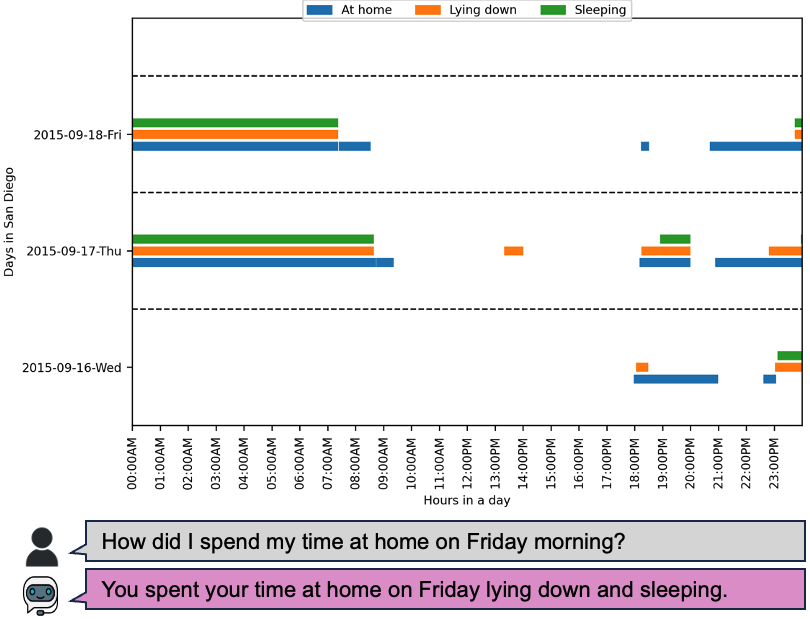}
        \vspace{-6mm}
        \caption{Example activity query.}
        \label{fig:weekly-activity}
    \end{subfigure}
    \vspace{-4mm}
    \caption{Example QA pairs in \Dataset. (a) and (b) are generated from daily graph, while (c) and (d) are generated from multi-day graph.}
    \label{fig:example_qas}
    \vspace{-4mm}
\end{figure*}

\vspace{-2mm}
\section{SensorQA Dataset}
\label{sec:dataset_creation}

In this section, we describe the collection process for the \Dataset dataset. The real-world scenario that \Dataset models is a long-term, in-the-wild sensor data collection scenario where users follow their daily routines without needing to focus on the sensing device. Throughout this process, users may pose arbitrary questions of personal interest about sensor data, whether focusing on a single day or over several weeks, and expect accurate answers derived from the collected data. To capture such real-world scenarios, we carefully design and implement protocols for both the sensor data (Sec.~\ref{sec:sensor-collect}) and the QA data (Sec.~\ref{sec:qa-collect}) collection for developing \Dataset.

\vspace{-2mm}
\subsection{Sensor Data Collection}
\label{sec:sensor-collect}


We choose to utilize a pre-existing dataset, the ExtraSensory dataset\footnote{The Extrasensory dataset is licensed under the Creative Commons Noncommercial License: \url{http://extrasensory.ucsd.edu/}} ~\cite{vaizman2017recognizing,vaizman2018context} as the source of the sensor data for \Dataset. We select ExtraSensory for its natural, in-the-wild collection setting and its massive scale. 
In contrast to the vast majority of sensor datasets~\cite{chavarriaga2013opportunity,ouyang2023harmony,yang2024mm} that are collected in a heavily controlled environment, ExtraSensory emphasizes real-life settings. This dataset uses easily accessible sensors (IMU, compass, location, audio, and phone state sensors on smartphones and smartwatches), with participants encouraged to maintain their natural routines throughout the collection period ranging from two days to three months. In contrast to Ego4D~\cite{grauman2022ego4d} which relies on data captured by a head-mounted camera, ExtraSensory imposes no restrictions on device placement, whether they be on desks, in pockets, or held in hand. These real-life collection protocols make ExtraSensory an ideal base for the \Dataset dataset.
Moreover, ExtraSensory contains raw sensor measurements from 60 subjects and has more than 300K minutes of data, tagged by 51 activity or context labels after cleaning.
The massive scale of ExtraSensory allows us to explore a wide range of real-life activities and personal routines during QA.
We also note that the data collection protocol for \Dataset can be extended to \textit{any} future sensor application.

\begin{table*}[htbp]
{
\small
\centering
\scalebox{0.83}{
\begin{tabular}{c|c|c}
\toprule
 Label Subset Name & Labels in Subset & Focus of Subset \\
\midrule
Main Activity & Lying down, Sitting, Standing, Walking, Bicycling & Posture pattern \\ \hline
Location & Indoor, Outside, At home, At main workplace, At school, In class & Location track \\ \hline
Dietary & Walking, Bicycling, Eating, Exercising & Eating and exercising patterns \\ \hline
Eat & Cooking, Eating, Cleaning & Eating patterns \\ \hline
Sleep & At home, Lying down, Sleeping & Sleeping patterns \\ \hline
Basic Needs & Sleeping, Eating, Grooming, Bathing, In Toilet & Physiological needs \\ \hline
Cleaning & Cleaning, Cooking, Grooming & Cleaning-related activities \\ \hline
Commute & In a vehicle, Walking, Outside, At home & Commute patterns\\ \hline
Exercise & Exercising, Walking, Bicycling & Exercising patterns \\ \hline
Electronics & Watching TV, Surfing the Internet, Phone in Hand, Computer Work & Activities using electronic devices \\  \hline
Social & Talking, With Friends, With Co-Workers & Activities with other people \\ \hline
Work & Sitting, Standing, Computer Work, Lab Work, In a meeting, In class, At school, At main workplace & Work-related activities and locations \\ \hline
Student & Computer Work, Lab Work, In class, At school & Study-related activities and locations \\ \hline
\multirow{2}{*}{Work-Life Balance} & Computer Work, Lab Work, In a meeting, At main workplace, At school, Watching TV, & \multirow{2}{*}{Work and relaxation patterns} \\
& Surfing the internet, At home  \\
\bottomrule
\end{tabular}
}
\caption{Label subsets used for organizing the graphs and encouraging focus on a specific aspect of daily life.}
\label{tab:label_subsets}
}
\vspace{-7mm}
\end{table*}

\vspace{-2mm}
\subsection{QA Data Collection}
\label{sec:qa-collect}

We use Amazon Mechanical Turk (AMT)~\cite{amt}, a crowdsourcing platform, to generate question-answer pairs through paid tasks completed by human workers. AMT is extensively utilized in the field of NLP for dataset creation~\cite{okvqav2,vqav2}. Unlike template-based question and answer generation in ~\cite{xing2021deepsqa} or simply using the narration text provided in the sensor dataset~\cite{moon2023anymal,han2024onellm}, our approach leverages the unique value of human-generated content, ensuring all QA pairs reflect genuine human interests and needs.

We carefully design our QA data collection process to ensure \textit{practical} and \textit{diverse} QA generation. Our objective is to produce a high-quality QA dataset that encompasses a variety of Q\&A types and time durations, thereby challenging the AI agent to perform effectively in real-world scenarios. To achieve this, we use two key strategies: (i) we develop multi-time scale activity graphs to facilitate the generation of both short-term and long-term questions, and (ii) we divide the context labels from the ExtraSensory dataset into subsets to encourage queries covering a wide range of aspects. Next, we detail each strategy.

\textbf{Creating multi-time scale activity graphs}
Collecting a dataset on AMT requires visualizing the sensor data for the workers and asking them to generate relevant Q\&A pairs. However, visualizing raw sensor signals presents a unique challenge due to the inherent unreadability of sensor data by humans~\cite{xing2021deepsqa}. Given our primary interest in understanding the underlying daily activities, we opt to visualize the activity or context labels over time and provide these graphs to the workers. These visual representations, depicted in Fig.~\ref{fig:example_qas}, resemble Gantt Charts, with the x-axis showing wall-clock time from 00:00 AM to 23:59 PM and the y-axis representing separate days. 
Different activity labels are shown by bars with distinct colors.
These graphs offer an intuitive visualization of daily activities along with the specific timestamps.

While the temporal scale of questions is determined by individual crowdsourcing workers, we recognize that the time scale depicted in activity graphs can implicitly influence their approach. For instance, when presented with a weekly activity graph, workers tend to ask more high-level and qualitative questions like "How frequently do I exercise?" rather than basic quantitative inquiries such as "What did I do at 10:00 AM?"
Motivated by this understanding, we have developed graphs at different temporal granularities to prompt questions across various scales:
\begin{itemize}[noitemsep,topsep=0pt]
    \item \textbf{Daily graph with timetable} We generate activity graphs for each user on a single day, accompanied by a table listing the start time, end time, and duration of all activities occurring on that day. This daily graph prompts workers to generate \textit{basic quantitative questions} about specific times and activities on a given day, as shown in Fig.~\ref{fig:daily-freq} and~\ref{fig:daily-time}, where we omit the detailed timetable due to space limitation.
    \item \textbf{Multi-day graph} We also create graphs depicting a user's activities over multiple days. The multi-day graph encourages workers to focus on the general and high-level activity patterns, leading to the generation of \textit{qualitative reasoning questions} as exemplified in Fig.~\ref{fig:weekly-day} and~\ref{fig:weekly-activity}.
\end{itemize}
Using multi-time scale activity graphs effectively balances the temporal scale of our collected dataset.

\textbf{Creating label subsets} 
The labels displayed in each activity graph guide workers to focus on specific aspects of daily life. Therefore, we carefully group these labels into subsets to cover comprehensive and diverse life aspects that users may want to monitor.
We create 14 subsets of labels as listed in Table~\ref{tab:label_subsets}, covering everything from essential living needs to work to social activities.
During \Dataset collection, we visualize one subset per graph and generate an equal number of questions per label subset.
This approach ensures the diversity of questions and answers, enhancing the practicality and difficulty of our dataset.

\textbf{QA data collection} We released the graphs and conducted the QA data collection on AMT over a three-week period. 
To maintain a balance between quantitative and qualitative questions, we requested for one question per daily graph and three questions per multi-day graph. 
To make the process more practical, workers were instructed to role-play as if the data were from their own wearable devices. They were asked to create questions from a first-person perspective, based on what would genuinely interest them, and to provide answers from a second-person perspective. The instructions were as follows: \textit{``Pretend you have a smart device with access to the information in the graph. Look at the graph and create a first-person question that requires information from the graph and would interest someone with a smart device. Then, answer in the second person, using the provided examples as a guide.''}
We then offered six QA examples, generated by the authors, for the workers to reference.

\vspace{-2mm}
\section{Dataset Analysis}
\label{sec:dataset_analysis}
In this section, we provide quantitative and qualitative analysis of the \Dataset to better understand its characteristics.

Examples of the collected Q\&As in \Dataset are displayed in Fig.~\ref{fig:example_qas}.
\Dataset contains 5,648 question-answer pairs, with an average length of 10.43 words per question and 10.48 words per answer. The dataset has a total of 118,051 tokens, of which 1,709 are unique and primarily related to daily activities. The repetition of words makes it more challenging for AI agents to answer questions accurately, as they must differentiate between similar questions based on the specifics of the sensor data.

To closely inspect the diversity of \Dataset, we profile the question and answer categories. We manually label 200 pairs, then train two BERT models~\cite{devlin2018bert} to classify the question and answer categories, respectively. 
The final profiling results are displayed in Table~\ref{tab:sensorqa_profile}.
\Dataset includes six distinct question categories and seven answer categories.
The distribution of questions and answers is imbalanced, with a notable focus on time-related aspects of activities, as seen in the high number of questions in the "Time Compare" and "Time Query" categories. This pattern aligns with practical user interests but has not been observed in previous QA datasets for human activities~\cite{xing2021deepsqa,moon2023anymal,moon-etal-2023-imu2clip}.
In addition to time-related queries, \Dataset covers a wide range of other aspects, including action, location, counting, and existence, demonstrating its diversity and practicality.

\begin{table}[pt]
\begin{subtable}[t]{0.48\textwidth}
\footnotesize
\centering
\begin{tabular}{c p{4.5cm} c} 
\toprule
\textbf{Question} & \textbf{Example Questions} & \textbf{\# of} \\
\textbf{Categories} & & \textbf{Questions} \\
\midrule
Time Compare & Did I spend more time sitting or standing? & 1,432 \\
Day Query & On which day did I spend the most time at home? & 1,277 \\
Time Query & How long was I in class and at school? & 1,119 \\
Counting & How often did I groom? & 725 \\
Existence & Did I have a meeting on Wednesday? & 668 \\
Action Query & What did I do after I left home on Tuesday? & 428 \\
\bottomrule
\end{tabular}
\vspace{-1mm}
\caption{\small Question categories in the \Dataset dataset.}
\label{tab:question_profile}
\end{subtable}
\begin{subtable}[t]{0.48\textwidth}
\footnotesize
\centering
\begin{tabular}{c p{4cm} c} 
\toprule
\textbf{Answer} & \textbf{Example Shortened Answers} & \textbf{\# of} \\
\textbf{Categories} & & \textbf{Answers} \\
\midrule
Action & Doing computer work & 1,357 \\
Day/Days & Last Friday & 1,242 \\
Existence & Yes/No & 1,047 \\
Time Length & 40 Minutes & 1,018 \\
Location & At school & 792 \\
Count & Three times & 401\\
Timestamp & Around 11:00 am & 310 \\
\bottomrule
\end{tabular}
\caption{\small Answer categories in the \Dataset dataset. We present a shortened version of the original answers for simplicity.}
\vspace{-2mm}
\label{tab:answer_profile}
\end{subtable}
\caption{\small Q\&A categories in \Dataset.}
\label{tab:sensorqa_profile}
\vspace{-7mm}
\end{table}



\vspace{-2mm}
\section{Benchmark Results}
\label{sec:baseline_and_results}

In this section, we benchmark state-of-the-art AI models on the \Dataset dataset and reveal the gap between existing models and ideal performance.

\vspace{-2mm}
\subsection{Benchmark Setup}

We establish comprehensive baselines using three distinct modality combinations: text-only, vision+text, and sensor+text, to identify the impact of each modality. 
We use few-shot learning (FSL) for closed-source models like GPT, and apply LoRA fine-tuning (FT)~\cite{hu2021lora} for open-source models like Llama. 
We randomly split 80\% of the data in \Dataset for training and 20\% for testing. 
More details are included in the github repository\footnote{\url{https://github.com/benjamin-reichman/SensorQA?tab=readme-ov-file}}.

\textbf{Baselines}
We begin by evaluating the generative pretrained models using few-shot QA examples in the prompt.
\begin{itemize}[topsep=0pt, itemsep=0pt]
    \item \textbf{GPT-3.5-Turbo~\cite{gpt-3.5} and GPT-4~\cite{gpt-4}} are \textcolor{mygreen}{text-only} baselines that only use the questions as input.
    \item \textbf{GPT-4-Turbo~\cite{gpt-4} and GPT-4o~\cite{gpt-4}} are \textcolor{myred}{vision+text} baselines that use both the questions and activity graphs.
    \item \textbf{IMU2CLIP-GPT4~\cite{moon-etal-2023-imu2clip}} is the state-of-the-art \textcolor{myblue}{sensor+text} GPT baseline. It first uses a trained CLIP model to retrieve the most relevant text for each chunk of IMU signal, then combines the text into a storyline and provides it to GPT-4, along with the question, for answer generation. We train the CLIP model using ExtraSensory~\cite{vaizman2017recognizing,vaizman2018context}.
\end{itemize}

We also selected the following open-source models, which are tested after they are either trained or finetuned. We started from the official code release and used the default hyperparameters. All Llama-based backbones are Llama-2 7B~\cite{touvron2023llama} unless specified otherwise.
\begin{itemize}[topsep=0pt, itemsep=0pt]
    \item \textbf{T5~\cite{2020t5}} and \textbf{Llama~\cite{touvron2023llama}} are widely used language models, serving as \textcolor{mygreen}{text-only} baselines.
    
    \item \textbf{Llama-Adapter~\cite{zhang2023llama}} is a \textcolor{myred}{vision+text} framework that combines vision inputs (activity graphs) with a Llama model via a pretrained transformer-based adapter.
    
    \item \textbf{Llava-1.5~\cite{liu2024improved}} is a state-of-the-art \textcolor{myred}{vision+text} model that integrates a visual encoder and Vicuna~\cite{vicuna2023} for visual and language understanding.
    
    \item \textbf{DeepSQA-CA~\cite{xing2021deepsqa}} fuses \textcolor{myblue}{sensor+text} modalities by training a CNN-LSTM model with compositional attention, for predicting from a limited set of answers given questions and IMU signals.
    
    \item \textbf{OneLLM~\cite{han2023onellm}} is a state-of-the-art multimodal LLM framework that supports eight modalities, including \textcolor{myblue}{sensor+text} data. It feeds sensor data through a pretrained CLIP encoder and uses a mixture of projection experts for modality alignment.
\end{itemize}


\textbf{Metrics} We consider two versions of \Dataset.
For the full-answer version, we use commonly applied NLP metrics, Rouge~\cite{lin2004rouge}, Meteor~\cite{banerjee2005meteor} and Bleu~\cite{papineni2002bleu} scores, to assess the semantic quality of answers.
We further distill a short-answer version of \Dataset by prompting GPT-3.5-Turbo~\cite{gpt-3.5} to extract the 1-2 keywords from each full answer. We use exact-match accuracy, i.e., whether the keywords appear in the generated answers, to evaluate the correctness of QA.

For efficiency, we aim for baseline models to ultimately be deployable on edge devices as personal AI assistants. Therefore, we evaluate their memory requirements and average generating latency per answer on the NVIDIA Jetson TX2~\cite{jetsontx2}, a typical edge platform featuring an NVIDIA Pascal GPU with 256 CUDA cores and 8GB of RAM. Here, all Llama-based models are quantized to 4-bit weights using AWQ~\cite{lin2023awq}.

\vspace{-2mm}
\subsection{Results}

\begin{table*}[htbp]
{
\small
\centering
\begin{tabular}{c|c|c|ccccc|c}
\toprule
 \textbf{Modalities} & \textbf{Backbone Model} & \textbf{FSL/FT$^1$} & \multicolumn{5}{c|}{\textbf{Full Answers}} & \textbf{Short Answers} \\
 & & & Rouge-1 ($\uparrow$) & Rouge-2 ($\uparrow$) & Rouge-L ($\uparrow$) & Meteor ($\uparrow$) & Bleu ($\uparrow$) & Accuracy ($\uparrow$) \\
\midrule
\textcolor{mygreen}{Text} & GPT-3.5-Turbo & FSL & 0.35 & 0.23 & 0.32 & 0.43 & 0.16 & 3.0\% \\
\textcolor{mygreen}{Text} & GPT-4 & FSL & 0.66 & 0.51 & 0.64 & 0.66 & \underline{0.39} & 16.0\% \\
\textcolor{mygreen}{Text} & T5-Base & FT & $0.71$ & $0.55$ & $0.69$ & \underline{$0.70$} & \textbf{0.43} & 25.4\% \\
\textcolor{mygreen}{Text} &  Llama & FT & \underline{$0.72$} & \textbf{0.62} & \textbf{0.72} & \textbf{0.72} & $0.38$ & 26.5\%  \\
\hline \hline
\textcolor{myred}{Vision+Text} & GPT-4-Turbo & FSL & 0.38 & 0.28 & 0.36 & 0.51 & 0.15 & 14.0\% \\
\textcolor{myred}{Vision+Text} & GPT-4o & FSL & 0.39 & 0.28 & 0.37 & 0.61 & 0.25 & 7.0\% \\
\textcolor{myred}{Vision+Text} & Llama-Adapter & FT & \textbf{0.73} & \underline{$0.57$} & \underline{$0.71$} & \textbf{0.72} & \textbf{0.43} & \textbf{28.0\%} \\
\textcolor{myred}{Vision+Text} & Llava-1.5 & FT & $0.62$ &  $0.46$ & $0.60$ & $0.58$ & $0.35$ & 21.5\%  \\
\hline
\hline
\textcolor{myblue}{Sensor+Text} & IMU2CLIP-GPT4 & FSL & 0.44 & 0.28 & 0.40 & 0.53 & 0.16 & 13.0\% \\ 
\textcolor{myblue}{Sensor+Text} & DeepSQA & FT & 0.34 & 0.05 & 0.34 & 0.18 & 0.0 & \underline{27.4\%} \\
\textcolor{myblue}{Sensor+Text} & OneLLM & FT & 0.12 & 0.04 & 0.12 & 0.04 & 0.0 & 5.0\% \\
\bottomrule
\multicolumn{9}{l}{$^{1}$\small{FS: Few-Shot Learning. FT: Finetuning.}} \\
\end{tabular}
}
\vspace{-1mm}
\caption{\small Benchmark results of baselines on \Dataset. Bold and underlined values show the best and second-best results.} 
\vspace{-7mm}
\label{tab:baseline_results_unfiltered}
\end{table*}

\textbf{QA performance} Table~\ref{tab:baseline_results_unfiltered} presents the results of all baselines on both the full-answer and short-answer versions of \Dataset. Rouge, Meteor, and Bleu scores assess n-gram precision between generated and reference answers, while exact-match accuracy focuses on correctness.

The results show that existing AI models perform poorly on \Dataset, with the best-performing baseline Llama-Adapter achieving an accuracy of only 28\%. This highlights the significant challenge \Dataset poses for current models, as \Dataset is designed to be realistic and diverse. Notably, the \textcolor{myblue}{Sensor+Text} baselines perform worse than even the \textcolor{mygreen}{Text-only} baselines, indicating that current models struggle with effective sensor-text fusion, particularly with \Dataset's long-duration sensor data. This fusion challenge seems to confuse the models, resulting in worse performance than using only the text of the question. These results emphasize the need for new approaches to effectively integrate sensor data and text in realistic applications like \Dataset.

Overall, finetuning open-source models outperforms few-shot learning on GPT baselines. For instance, Llama-Adapter achieves 0\% exact-match accuracy without fine-tuning, highlighting the importance of fine-tuning on the target dataset to better adapt models for specific tasks.

\begin{figure}[t]
\begin{center}
\begin{tabular}{cc}
    \vspace{-1mm}
    \includegraphics[width=0.22\textwidth, height=3cm]{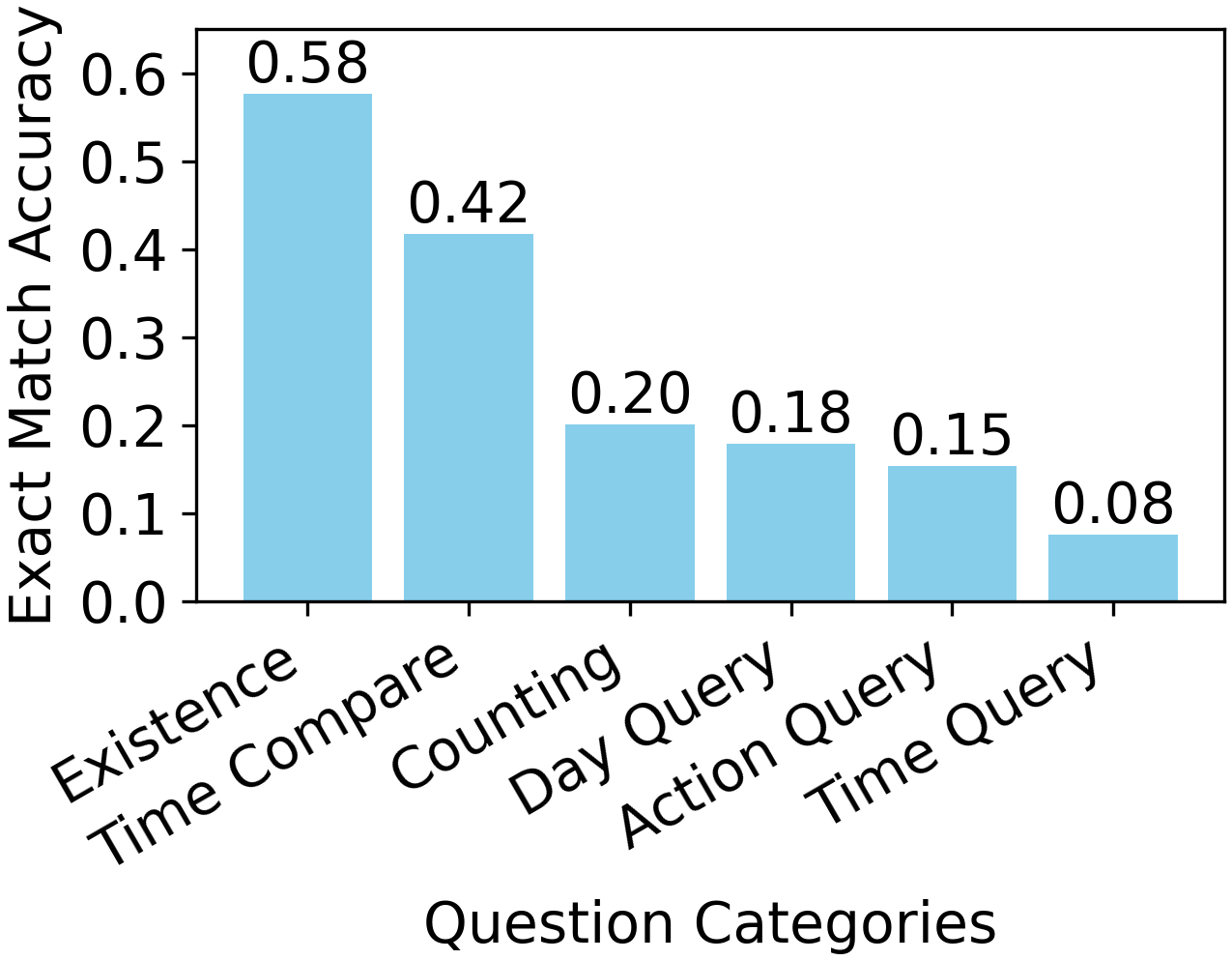} 
    &
    \includegraphics[width=0.22\textwidth, height=3cm]{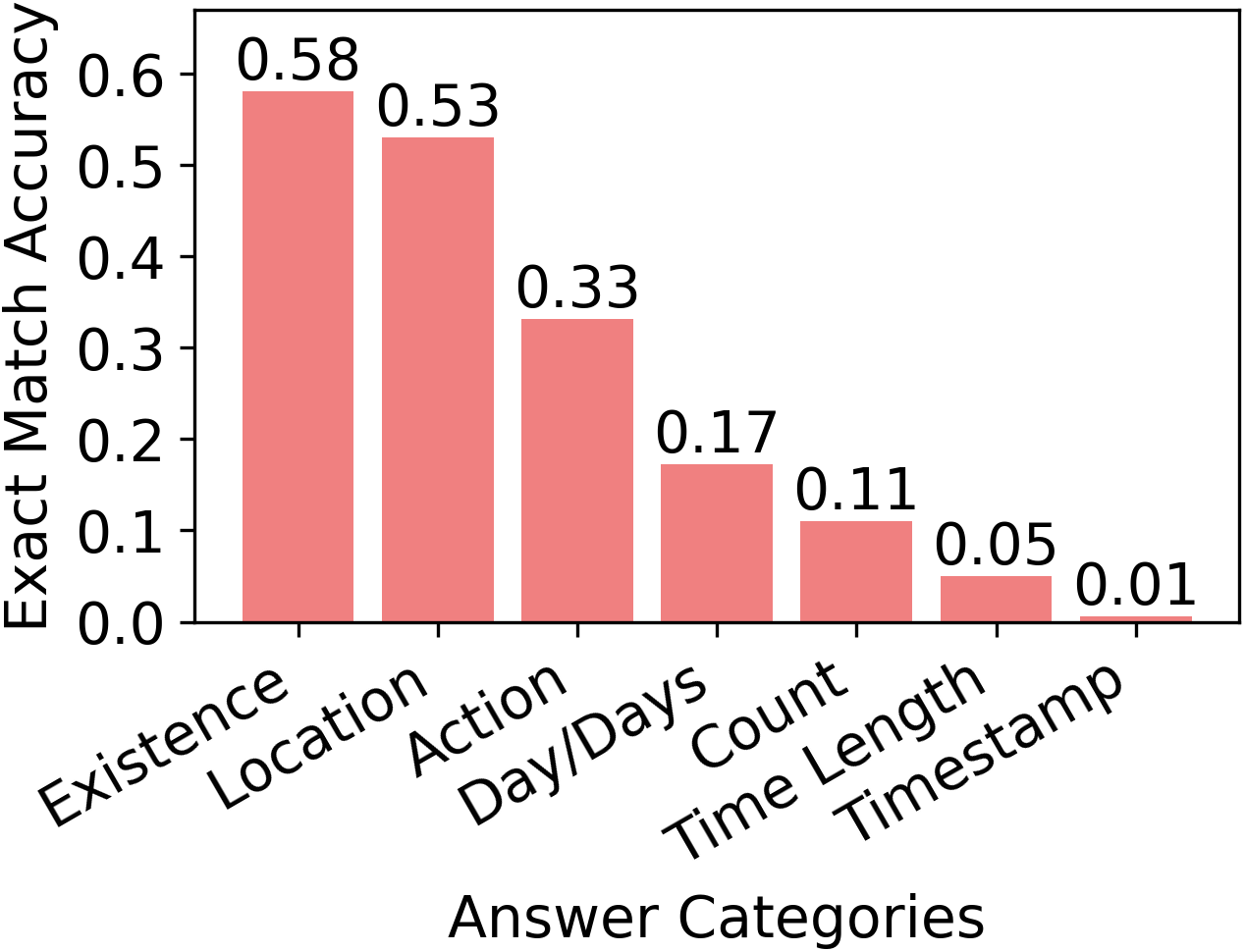}
\end{tabular}
\vspace{-5mm}
\caption{\small Exact-match accuracy of Llama~\cite{touvron2023llama} displayed by question category (left) and answer category (right).}
\label{fig:variants}
\end{center}
\vspace{-6mm}
\end{figure}

\textbf{QA performance per category} Different Q\&A types present varying difficulties for the models. Fig~\ref{fig:variants} shows exact-match accuracy by category using Llama~\cite{touvron2023llama}, one of the strongest baselines. Existence questions achieve the highest accuracy since they require only a Yes/No response. However, Llama performs only slightly better than random guessing, with an accuracy of 58\%. The most challenging category involves time-related queries, such as duration and timestamp questions, which makes \Dataset unique compared to prior datasets~\cite{xing2021deepsqa,moon2023anymal,moon-etal-2023-imu2clip}. It is critical for future approaches to accurately extract time information from sensor data and incorporate it into responses.

\begin{figure}[t]
\begin{center}
\includegraphics[width=0.43\textwidth, height=3cm]{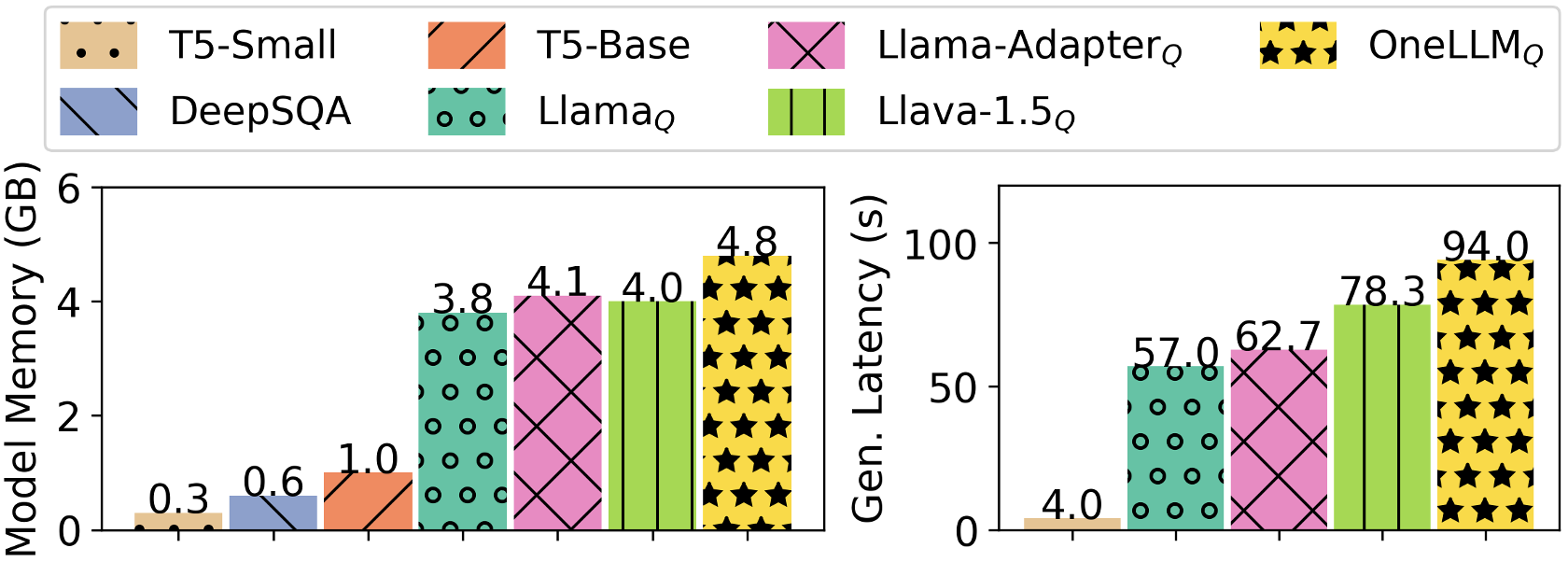} 
\vspace{-5mm}
\caption{\small Model memory size (left) and average answer generating latency (right) on Jetson TX2~\cite{jetsontx2}. Footnote $Q$ denotes models after quantization.}
\label{fig:eff}
\end{center}
\vspace{-6mm}
\end{figure}

\textbf{Efficiency results} Fig.~\ref{fig:eff} presents model memory requirements and average answer generation latency on the NVIDIA Jetson TX2~\cite{jetsontx2}. Non-LLM models generally require less memory; however, DeepSQA and T5-Base encounter out-of-memory (OOM) issues with large multi-day timeseries inputs, so their latency results are omitted. Non-LLM methods also show poor QA performance.
LLM-based models, while more accurate, demand substantial memory for their billion-level parameters, resulting in an average generation latency of over 57 seconds, which is impractical. Optimizations in memory and efficiency are essential to develop a viable conversational AI agent for edge deployment. We hope \Dataset encourages new contributions in this area.

\vspace{-2mm}
\section{Conclusion}
\label{sec:conclusion}
As IoT and wearable devices proliferate, the ability to interact with sensor data through conversational AI becomes increasingly crucial.
In this work, we introduce SensorQA, a question-answering dataset created by humans to foster natural language interactions between humans and wearable sensors in daily life monitoring. \Dataset is built on sensor data from ExtraSensory~\cite{vaizman2017recognizing,vaizman2018context} and 5.6K QA pairs collected from AMT, featuring practical scenarios and diverse queries.
Benchmarking results on state-of-the-art AI models demonstrate the gap between existing solutions and ideal performance, both in QA and efficiency.



\begin{acks}
This work was supported in part by National Science Foundation under Grants \#2003279, \#1826967, \#2100237, \#2112167, \#1911095, \#2112665, \#2120019, \#2211386 and in part by PRISM and CoCoSys, centers in JUMP 2.0, an SRC program sponsored by DARPA.
\end{acks}
\bibliographystyle{ACM-Reference-Format}
\bibliography{main}


\end{document}